\documentclass[acmtog,authorversion]{acmart} 

\usepackage{booktabs} %

\usepackage[ruled]{algorithm2e} %

\usepackage{dsfont}

\SetAlFnt{\small}
\SetAlCapFnt{\small}
\SetAlCapNameFnt{\small}
\SetAlCapHSkip{0pt}

\acmJournal{TOG}

\begin{document}

\title{Fovea Stacking: Imaging with Dynamic Localized Aberration Correction}

\author{Shi Mao}
\orcid{0000-0001-9275-4632}
\affiliation{
 \institution{King Abdullah University of Science and Technology (KAUST)}
 \city{Thuwal}
 \country{Saudi Arabia}}
\email{shi.mao@kaust.edu.sa}

\author{Yogeshwar Nath Mishra}
\authornote{Now at Indian Institute of Technology Jodhpur}
\orcid{0000-0003-2063-2200}
\affiliation{
 \institution{King Abdullah University of Science and Technology (KAUST)}
 \city{Thuwal}
 \country{Saudi Arabia}}
\email{mishrayn@iitj.ac.in}

\author{Wolfgang Heidrich}
\orcid{0000-0002-4227-8508}
\affiliation{
 \institution{King Abdullah University of Science and Technology (KAUST)}
 \city{Thuwal}
 \country{Saudi Arabia}}
\email{wolfgang.heidrich@kaust.edu.sa}

\renewcommand\shortauthors{Mao, S. et al}

\begin{abstract}
The desire for cameras with smaller form factors has recently led to a push for exploring computational imaging systems with reduced optical complexity such as a smaller number of lens elements. Unfortunately such simplified optical systems usually  suffer from severe aberrations, especially in off-axis regions,  which can be difficult to correct purely in software.

In this paper we introduce Fovea Stacking, a new type of imaging system that utilizes an emerging dynamic optical component called the deformable phase plate (DPP) for localized aberration correction anywhere on the image sensor.  By optimizing DPP deformations through a differentiable optical model, off-axis aberrations are corrected locally, producing a foveated image with enhanced sharpness at the fixation point - analogous to the eye’s fovea. Stacking multiple such foveated images, each with a different fixation point, yields a composite image free from aberrations. To efficiently cover the entire field of view, we propose joint optimization of DPP deformations under imaging budget constraints. Due to the DPP device's non-linear behavior, we introduce a neural network-based control model for improved agreement between simulation and hardware performance.

We further demonstrated that for extended depth-of-field imaging, Fovea Stacking outperforms traditional focus stacking in image quality. By integrating object detection or eye-tracking, the system can dynamically adjust the lens to track the object of interest-enabling real-time foveated video suitable for downstream applications such as surveillance or foveated virtual reality displays.

\end{abstract}



\begin{teaserfigure}
  \includegraphics[width=\textwidth]{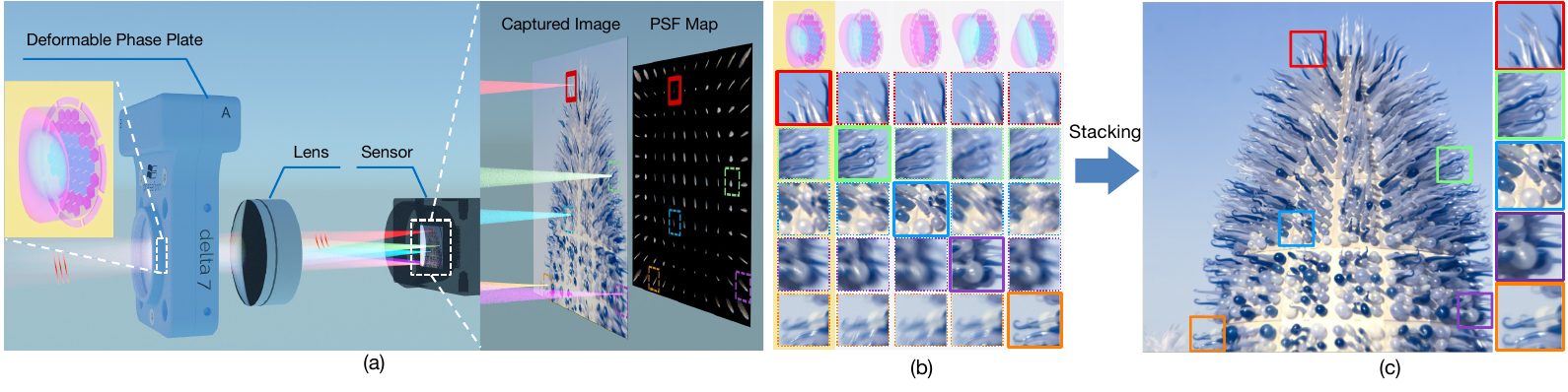}
  \caption{
    Fovea Stacking is a novel computational camera paradigm aimed at simplifying (and ultimately miniaturizing) camera optics. (a) The optical system consists of an achromatic doublet lens, which is highly aberrated, and a refractive deformable phase plate (DPP), which can be used to correct the aberrations for a {\em local} region of interest (the fovea). While this optical system cannot simultaneously correct all aberrations across the entire image, the DPP is capable of dynamically moving the fovea {\em anywhere} in the image. For example, the DPP deformation in (a) corrects aberrations along the red beam for the corresponding red box fovea on the sensor. (b) shows magnified crops of foveated images captured with different DPP deformations. The left column shows the image from (a), the other columns have the fovea moved to the other respective boxes. (c) By stacking up the sharpest regions from different foveated images (i.e. a fovea stack), an un-aberrated all-in-focus image can be recovered. Other applications such as object tracking and depth-dependent focusing are discussed in the text.
  }
  \Description{presenting the general idea}
  \label{fig:teaser}
\end{teaserfigure}

\maketitle

\section{Introduction}
To achieve high image quality across the full field of view (FoV), conventional imaging systems use complex lens systems that are carefully designed to correct for optical aberrations both on- and off-axis. Great strides have been taken to miniaturize such designs to achieve impressive imaging results in small form factors, for example in mobile devices. Unfortunately further miniaturization is approaching physical limits with existing design philosophies and optical components such as refractive lenses.

New opportunities arise from the emergence of new types of dynamically tunable optical components that facility completely new design paradigms. The most well-known type of such elements are probably liquid tunable lenses~\cite{liu2023FTL}. Initially bulky and expensive, this technology has recently become small and affordable enough to be used in mobile phones~\cite{Blain2021xiaomi}. %
A more recent type of dynamically tunable optical component is the deformable phase plate (DPP)~\cite{banerjee2018optofluidic,rajaeipour2021seventh}. Similar to a liquid tunable lens, DPPs allow for dynamic control of a liquid surface shape for optical purposes. However, instead of only controlling a global lens curvature, DPPs provide fine control over local surface geometry, thereby offering a transmissive alternative to the reflective
deformable mirrors used in adaptive optics systems.

\begin{figure}
    \centering
    \includegraphics[width=\linewidth]{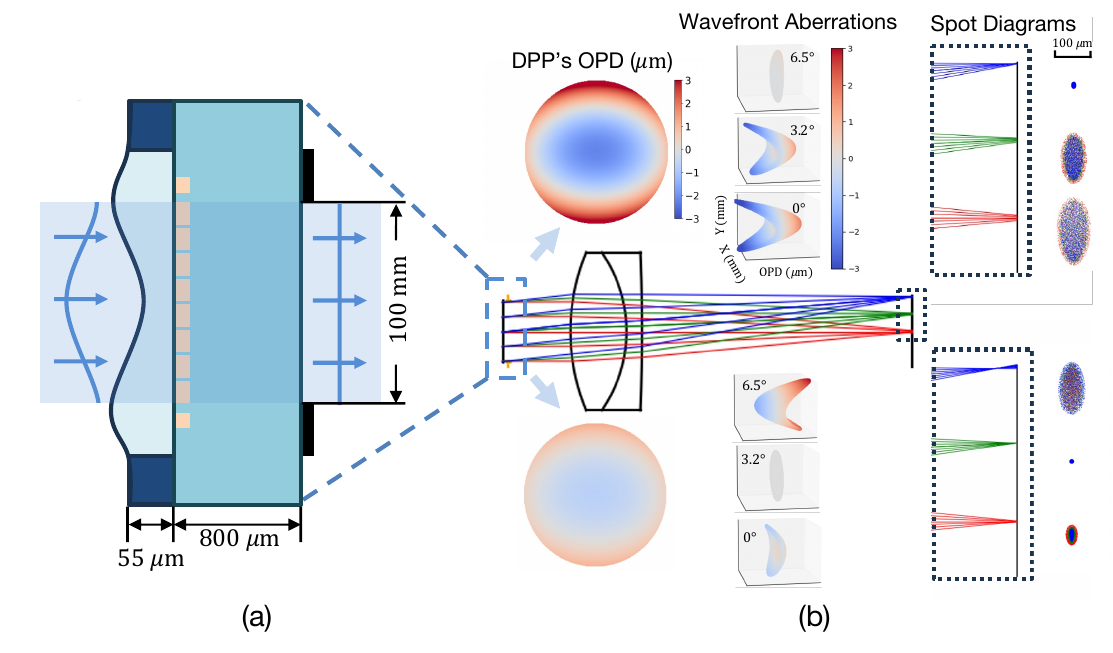}
    \caption{  (a) physical layout of DPP device, composed of a deformable membrane, a optical liquid cavity, and a rigid glass substrate containing hexagonal electrodes. Figure adapted from \cite{rajaeipour2021seventh}. (b) localized aberration of different oblique angles can be corrected by dynamically changing DPP patterns. The green-channel wavefront aberration on exit-pupil plane and the corresponding colored spot diagrams are shown.
    }
    \label{fig:dpp_physical_principle}
\end{figure}

In this work we use the ability of the DPP to {\em dynamically} and {\em locally} shape an optical wavefront in a compact optical system to demonstrate a new type of foveated imaging system (see Fig.~\ref{fig:teaser}). The optical design consists of a combination of a highly aberrated achromatic double lens and a DPP that can be used for
aberration correction. This optical system lacks the complexity to
{\em simultaneously} correct for {\em all} aberrations {\em everywhere} in the image like a classical camera lens, however, it has the ability to correct for the aberrations in a localized region of interest (the fovea) and produce an image with excellent fidelity in that region. Moreover, through dynamic control of the DPP, this fovea region can be placed anywhere in the image plane using direct electrostatic actuation (c.f. Fig.~\ref{fig:teaser}b, Fig.~\ref{fig:dpp_physical_principle}). A stack of such foveated images, a.k.a. a {\em Fovea Stack}, can be used to reconstruct an unabberrated, high fidelity image (c.f. Fig.~\ref{fig:teaser}c).

Specifically, we make the following contributions in this work:
\begin{itemize}
\item we propose Fovea Stacking as a new paradigm for camera systems;   we optimize the DPP's wavefront control patterns for various depths
  using differentiable optics, resulting in regionally corrected images that can be stacked to produce an aberration-free composite image.
\item to efficiently cover the FoV with minimal saccades, we propose a method that jointly optimizes the DPP's deformation patterns, allowing effective stacking of just 3-5 images for full aberration correction.
\item to accurately model non-linear DPP behavior for larger control signals, we develop a neural network model for mapping control signals to actual wavefront shapes and utilize this model to reduce the gap between simulation and real experiments.
\item we demonstrate extended depth of field imaging by Fovea Stacking across a range of different depths.
\item by integrating foveate imaging with object detection or eye tracking, we also demonstrate smooth pursuit movement that dynamically adjusts the imaging system to keep an object of interest within the fovea region, enabling the generation of foveated video suitable for downstream tasks such as surveillance or foveated VR display.
\item we demonstrate all the above capabilities on a hardware prototype that is already compact enough to be used outside the lab, although DPPs have not yet reached the same level of miniaturization as liquid tunable lenses.
\end{itemize}

\def\vec{\mathbf}

\section{Related Work}

\subsection{Wavefront Modulators} 

Wavefront modulators are optical devices capable of dynamically shaping light to compensate for optical distortions. Deformable mirrors (DMs)~\cite{bifano2011mems}, which are typically reflective, require folded optical paths. Liquid crystal phase-only spatial light modulators (LC SLMs)~\cite{yang2023LCSLM}, are frequently used for wavefront shaping in microscopy or holography however modern variants are usually reflective designs, since the pixel wiring introduces diffraction artifacts in older transmissive designs. Either variant requires polarized light and exhibits a strong wavelength dependency.

Recently, a novel optofluidic refractive wavefront modulator named a deformable phase plate (DPP) was proposed by \cite{banerjee2018optofluidic}. 
As shown in Fig.~\ref{fig:dpp_physical_principle}a, the DPP consists of a thin optically transparent, liquid-filled cavity sandwiched between a deformable membrane and a rigid glass surface. Its shape can be controlled by exerting electrostatic forces on the membrane via voltages applied to a hexagonal grid of 63 transparent electrodes on the rigid glass back panel. Using this principle, a DPP enables distortion correction up to seventh radial Zernike order~\cite{rajaeipour2021seventh}.
In this work, we adopt the DPP as a refractive wavefront modulator to create a compact imaging system. However, it is observed that traditional linear control strategies are inaccurate for large wavefront deviations. To address this, we propose a neural network-based control strategy to improve accuracy.

Compared to adaptive optics (AO) systems~\cite{hampson2021adaptiveoptics,wang2018megapixelAO}, which typically correct for aberrations introduced by atmospheric turbulence or other external media before light reaches the imaging system, our method focuses on correcting distortions caused by the inherent imperfections of simple, compact imaging optics, tailored to the specific optical system in use.

\subsection{Focus Stacking}

Focus Stacking extends depth of field by capturing multiple images at varying focus distances. From this image stack, a sharp all-in-focus (AIF) image can be reconstructed. Similarly, Fovea Stacking aims to reconstruct a sharp, aberration corrected image from a stack of regionally sharp images.
Focus Stacking can also be used correct for field curvature, where the
focal region is a curved manifold instead of a plane. It can therefore
serve as a baseline for our work.

Traditional multi-focus image fusion methods reconstruct all-in-focus images by identifying and compositing the sharpest regions from multiple input images ~\cite{huang2007evalFocusMeasure,nayar1994shape}. This fusion can occur in the spatial domain or in transformed domains, such as Laplacian pyramids~\cite{wang2011LaplacianPyramid,sun2018multiLaplacianPyramid} and discrete wavelet transforms~\cite{pradnya2013wavelet}. For comprehensive reviews of traditional and deep learning-based methods, see~\cite{zafar2020MFIF_Review,zhang2021DL_MFIF_Review}. Alternatively, some approaches jointly reconstruct both a depth map and an all-in-focus image - for example, Wang et al.~\shortcite{wang2021AIFDepthNet} use 3D convolution to learn attention weights from the focus stack, which are then used to produce both the AIF image and the corresponding depth map.

Traditional focus adjustment often involves mechanical movement of the sensor or lens elements, which can induce vibrations during large-amplitude, high-frequency motion~\cite{kuthirummal2010flexible}. As an alternative, focus-tunable liquid lens (FTLs) have been adopted for focus-sweeping applications~\cite{miau2013focal}, but they are limited to tuning optical power without the ability to correct higher-order aberrations. Although the adjustable range of the DPP is limited, it  is capable of adjusting the focus distance by modulating the "defocus" term in the Zernike polynomial. Our method, therefore, supports stacking not only in the x-y direction but also along the depth direction, enabling aberration-corrected imaging for extended depth-of-field applications. 

\subsection{Differentiable Optics}

Recent advances in computational optics demonstrate the effectiveness of end-to-end co-design of optics and reconstruction algorithms~\cite{sitzmann2018end,sun2021complexLens}. Central to this approach is a differentiable optics model. The first differentiable implementation of light propagation through a diffractive optical element (DOE)~\cite{sitzmann2018end} enabled applications in extended depth of field and super-resolution. This DOE model has since supported optical designs for wide field-of-view imaging~\cite{peng2019learned}, coded super-resolution SPAD imaging~\cite{sun2020end}, and multi-aperture hyperspectral imaging~\cite{shi2024learned}. To support the end-to-end design of compound refractive lenses beyond simple wave-optics systems, researchers introduced both neural proxies~\cite{tseng2021differentiable} and differentiable ray-tracing~\cite{sun2021complexLens}, which computes ray–surface intersections and refractions with full differentiability with respect to lens parameters. Subsequent work on differentiable ray-tracing reduced memory usage~\cite{wang2022dO} and enabled automatic lens design from random initialization~\cite{yang2024curriculum}, leading to the design of compound fluidic-freeform lenses~\cite{na2024fludicLenses} and even hybrid refractive–diffractive lenses~\cite{yang2024hybrid}.

In this work, we adopt a differentiable ray-tracing framework \cite{yang2024curriculum} to optimize DPP deformation patterns for foveated imaging and stacking, targeting reconstruction quality. The customized differentiable model for the DPP device is detailed in Sec.~\ref{sec:wavefront_optimization}.

\subsection{Foveated Imaging}

Foveated imaging, inspired by human vision, concentrates high
resolution where it is most needed - similar to the function of the fovea. Many different approaches have been proposed to achieve spatially varying resolution through specialized sensor design or image fusion techniques. Retina-like sensors, for
example, arrange photoreceptors in a spatially varying log-polar structure~\cite{sandini2003retina,boluda1996fovea_sensor}. Imaging systems using prism arrays ~\cite{carles2016multi_prism_array} or a Risley-prism \cite{huang2021flexible_Risley_prism} expand the FoV by deflecting incoming rays and digitally super-resolving the overlapping central region, effectively producing foveated images through fusion. Similarly, a 3D-printed microlens system
~\cite{thiele20173d_eagle_eye} combines pixels from four microlenses with different focal lengths to generate a foveated image. However, these methods hard-code the fovea region, necessitating mechanical camera motion to perform saccades.

More directly related to our work is a small body of literature on creating foveated imaging systems by locally correcting optical
aberrations using dynamic wavefront shaping methods~\cite{martinez2001foveated}.  Such foveated images result from localized aberration correction, which can effectively address only one limited region of interest at a time. To correct off-axis aberrations locally, various optical solutions using different
wavefront modulators have been proposed, including reflective SLMs~\cite{wick2002foveated_SLM,curatu2005wideFoV_SLM}, transmissive 
SLMs~\cite{harriman2006transmissive_SLM}, and reflective DMs~\cite{zhao2008wide_DM,zhao2008broadband_DM}. However, these methods are either bandwidth-limited and require polarized light (as with SLMs), or they require folded optical paths and complex system designs (as with reflective modulators), which is counterproductive for miniaturizing and simplifying optical systems. Furthermore, they typically optimize the optics for focusing at infinity and at a fixed
oblique angle. In this work, we propose the use of a refractive wavefront modulator to enable compact and wide-band imaging. Additionally, we introduce a differential optics-based optimization method to tailor the desired wavefront across different imaging distances.

\section{Wavefront Optimization with Differentiable Optics}
\label{sec:wavefront_optimization}
In this section, we describe how to optimize DPP wavefront control patterns to (1) dynamically position the fovea anywhere within the image and (2) efficiently cover the field of view (FoV) with minimal saccades. We develop a forward model of the optical system shown in Fig.~\ref{fig:teaser}, comprising an achromatic doublet lens and a deformable phase plate, to enable ray tracing through the system. This model is implemented in a differentiable manner, allowing us to optimize the DPP’s wavefront parameters via back-propagation of the imaging loss. The subsections are organized as follows: Sec.~\ref{sec:sub_DPP_Model} details the DPP model within differentiable optics system; Sec.~\ref{sec:sub_single_optimize} covers the optimization of a single DPP wavefront for foveated imaging at oblique angles. Sec.~\ref{sec:sub_joint_optimize} presents strategies for jointly optimizing multiple DPP wavefronts to cover the FoV for Fovea Stacking, under a limited number of images.

\subsection{DPP Surface Modeling}
\label{sec:sub_DPP_Model}
With an optofluidic cavity thickness of $55\mu m$, the DPP can be treated as a thin plate capable of phase modulation. Based on Fermat's principle, the generalized Snell's law of refraction~\cite{yu2011light} governs the direction of light propagation in response to such abrupt phase changes. Accordingly, in the differentiable optics system, we model the DPP as a refractive plane that bends light according to the optical path difference (OPD) it introduces.

\begin{align*}
    \alpha' =& \alpha + \frac{\partial D(x,y)}{\partial x}\\
    \beta' =& \beta + \frac{\partial D(x,y)}{\partial y} \\
    \gamma' =& \sqrt{1 - \alpha'^2 - \beta'^2} 
\end{align*}

Here, $\vec{n'} = (\alpha',\beta',\gamma')^T$ and $\vec{n} = (\alpha,\beta,\gamma)^T$  are the normalized directions of in-going and refracted rays,  respectively, and $D$ denotes OPD.
To represent the wavefront generated by the DPP, we parameterize the OPD as Zernike polynomials up to K polynomials:
\begin{equation}
    D(\rho,\varphi) = \sum_k^K w_k Z_k(\rho,\varphi),
\end{equation}
where $\rho$ is normalized radial distance (normalized by aperture radius), $\varphi$ is the azimuthal angle, $Z_k$  and $w_k$ are the k-th Zernike polynomials and corresponding coefficients following OSA indices~\cite{thibos2002OSA_standard}. Partial derivatives with respect to $x$ and $y$ can be therefore derived and parameterized by $w_k$. Because horizontal and vertical tilt only cause geometric distortion -- resulting in unwanted pixel misalignment across images rather than improved image sharpness -- they are excluded from Zernike coefficient optimization and fixed at zero. Notice that we do not correct for radial lens distortion since it remains consistent across a stack of images as long as the DPP does not introduce tilt.  More details on Zernike polynomials and their derivative calculation is provided in the Supplement.

\subsection{Optimized Wavefronts for Foveated Imaging}
\label{sec:sub_single_optimize}

To maximize the imaging quality at a specific oblique angle, corresponding rays are sampled and traced through the optical system, consisting of a DPP and a lens. The RMS spot size serves as the optimization loss, guiding the optimization of the DPP wavefront pattern via backpropagation.
Specifically, for each sampled object point $P$ at distance $D$, $N_a$ points $\{P_i | 1\leq i \leq N_a\}$ are sampled on the aperture plane to generate the sampling rays $\{R_i = Ray(P \rightarrow P_i)| 1\leq i \leq N_a\}$. Each sampled ray is then traced through the optical stack to reach the image plane at positions $p_{i\lambda} = RT(R_i;\lambda)$, given a specific wavelength $\lambda$. The root-mean-square radial deviation of traced rays from their center, termed the \textbf{RMS spot size}, is used as the loss function minimized via backpropagation. As RMS spot size is very closely related to the area under the modulation transfer function MTF~\cite{luc2022aberration}, it is a popular loss in lens design. For color images, the RMS spot size is further integrated over multiple wavelengths.

\begin{equation}
    r(P) = \frac{1}{3} \sum_{\lambda}^{3} \frac{1}{N_a}\sum_i^{N_a}{||p_{i\lambda} - \Bar{p}||_2}
\end{equation}

\begin{figure}
    \centering
    \includegraphics[width=0.8\linewidth]{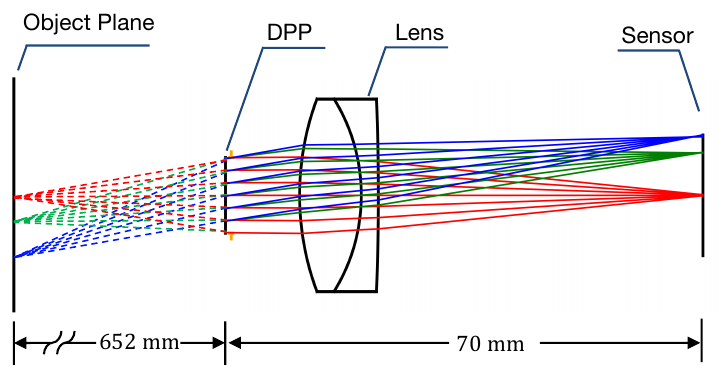}
    \caption{Side view of the calibrated simulation system. The object plane is set 652 mm away. The imaging setup includes a DPP, a 50 mm achromatic doublet lens (Thorlabs AC-254-050A), and a Bayer pattern RGB sensor (FLIR GS3-U3-41C6C-C).}
    \label{fig:simulation_system}
\end{figure}

\begin{figure}
    \centering
    \includegraphics[width=\linewidth]{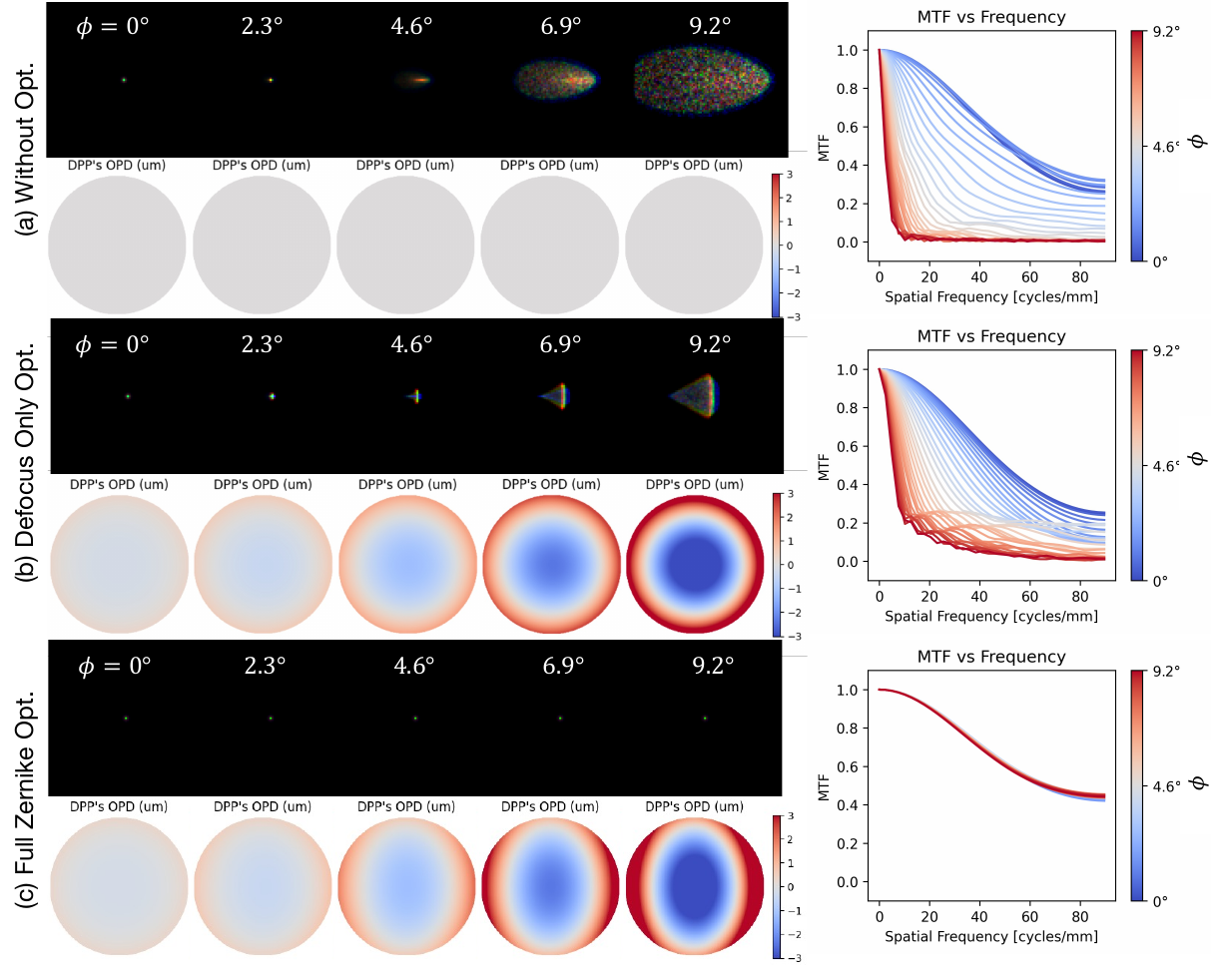}
    \caption{Optimized PSF and MTF for various oblique angles $\phi$ relative to the optical axis: (a) initial system without DPP, (b) optimization of the defocus Zernike parameter only, simulating improvements that can be made by focus adjustments (either by shifting the lens or with a liquid tunable lens) and (c) full Zernike optimization up to 4th order. MTF values are averaged over sagittal and tangential components. }
    \label{fig:MTF-FoV}
\end{figure}

To enhance convergence, we reduce the learning rate for each Zernike coefficient by a factor of  $\sqrt{10}$ for each successive polynomial order. Since the horizontal/vertical tilting terms only shift the image and do not introduce optical degradation, they are excluded to prevent ray deflection.

To verify the DPP's capability for localized aberration correction, an optical system is set up and calibrated in a real-world setting as described in Sec.~\ref{sec:calibration}, and shown in Fig~\ref{fig:simulation_system}. A total of 32 oblique angles $\phi$ are sampled within the range of $9.2^\circ$ (corresponding to half the sensor's diagonal size $R$) along the horizontal axis. The PSFs for five angles of these angles are visualized in Fig~\ref{fig:MTF-FoV}. The average modulation transfer function (MTF) is computed from the PSFs by first performing a weighted average of the RGB channels based on relative luminance, followed by averaging the sagittal and tangential MTFs. The initial setting without any optimization, exhibits severe off-axis aberrations (Fig~\ref{fig:MTF-FoV}a). While adjusting the focus - equivalent to optimizing only the defocus term in Zernike polynomials - can mitigated field curvature (Fig~\ref{fig:MTF-FoV}b), other aberrations remain uncorrected. Full optimization up to the 4th order of Zernike polynomials effectively corrects these off-axis aberrations (Fig~\ref{fig:MTF-FoV}c).

\begin{figure}
    \centering
    \includegraphics[width=\linewidth]{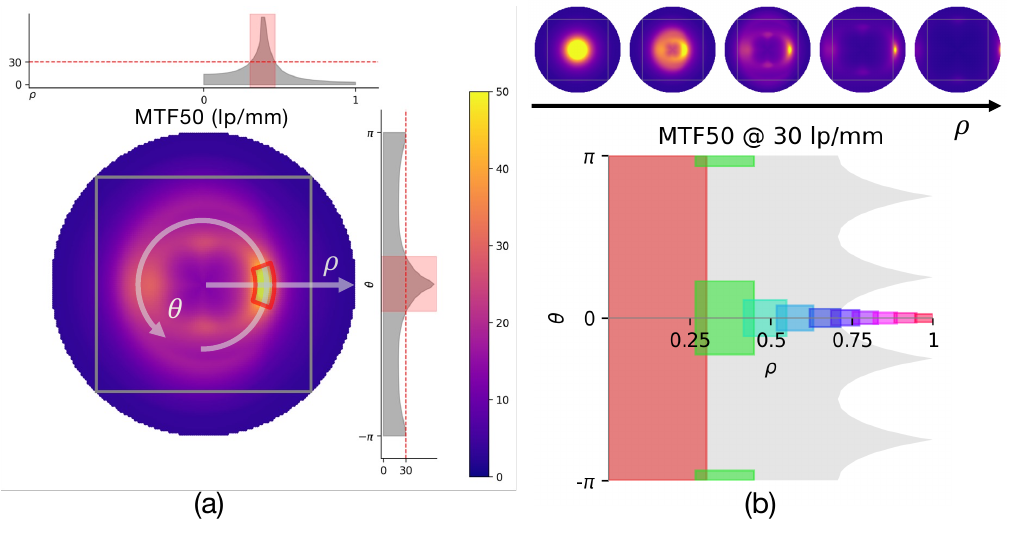}
    \caption{Radial and angular coverage of the fovea area. (a) For a single optimized DPP pattern, the fovea area—defined as the area where MTF50 exceeds a threshold like 30 lp/mm—can be characterized by its extent along the normalized radial axis $\rho$ and angular axis $\theta$. The gray square indicates the sensor region. (b) As the optimized angle approaches the sensor's edge, the fovea area shrinks, necessitating more images in the peripheral areas to maintain sharpness over the gray area representing the sensor.}
    \label{fig:MTF50-FoV}
\end{figure}

Although aberration can be effectively corrected locally for different oblique angles $\phi$, the size of the fovea varies by location. In Fig~\ref{fig:MTF50-FoV}, for each optimized DPP wavefront pattern, the MTF50 (defined as the spatial frequency at which the MTF drops to $50\%$) is visualized within a circular FoV with radius $R$, which circumscribes the square sensor (shown as a gray square). Given a specific sharpness threshold (e.g., 30 line pairs per mm), the radial and angular regions exceeding this threshold are identified and plotted as a rectangular region in a normalized radial-angular diagram. The normalized radius $\rho$ is related to the angle $\phi$ by: $\rho=f \tan(\phi)/R$, where $f$ is the effective focal length. The sharp coverage area tends to shrink as it approaches the periphery. It is estimated that approximately 100 images are needed to cover the entire square sensor region (shaded in gray in the radial-angular diagram) with resolution exceeding 30 lp/mm. More detailed analysis is provided in Supplementary. To capture sharp images across the full FoV more efficiently, a trade-off between sharpness and coverage must be considered. Notably, a secondary peak in sharpness often appears in the opposite direction (as shown at $\theta = \pm\pi$ in Fig.~\ref{fig:MTF50-FoV}a), due to the known symmetries of certain aberrations, which can potentially be exploited.

\begin{figure*}
    \centering
    \includegraphics[width=\linewidth]{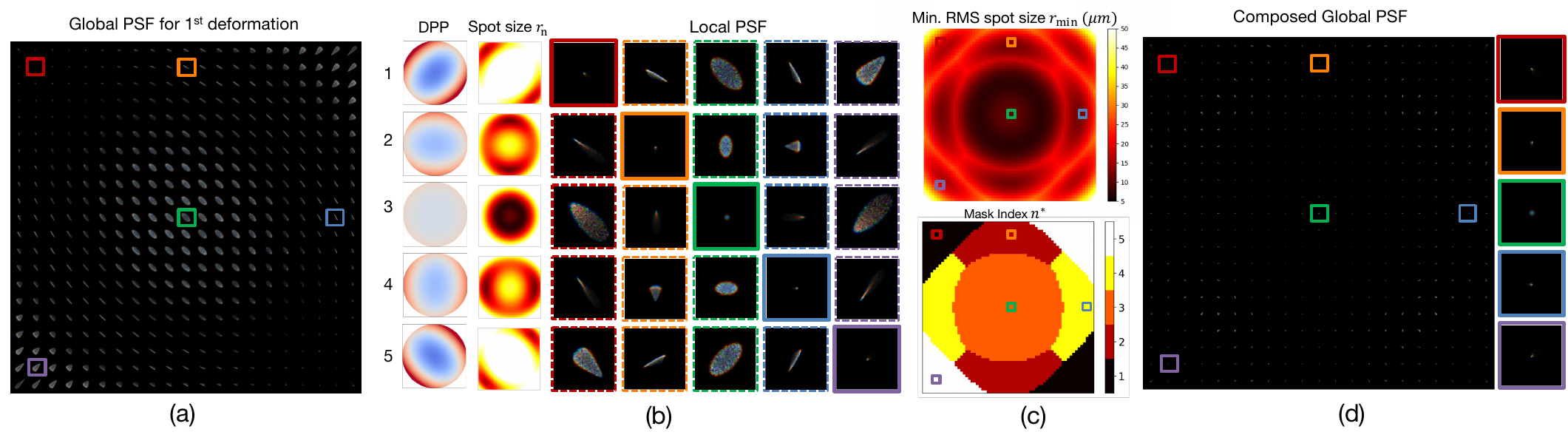}
    \caption{Joint optimization of 5 DPP patterns using grid stacking. (b) After optimization, each DPP deformation is specialized to optimize quality over different local regions while jointly covering the full FoV. (a) As an example the $1^{st}$ DPP pattern produces the best focus in the top-left and bottom-right regions. (c) The minimal value of RMS spot size $r_{min}$ across the stack serves as the optimization loss, while the mask index $n^*$ indicates which DPP pattern is best for any given region, which can be used to generate (d) a composite PSF map for the image after stacking. }
    \label{fig:psf_grid}
\end{figure*}

\subsection{Optimized Wavefronts for Fovea Stacking}
\label{sec:sub_joint_optimize}

\subsubsection{ROI tiling}

Aberration correction across the field of view can be achieved by dividing the FoV into regions of interest (ROI) and correcting local aberrations independently. To optimize each wavefront for local correction within a ROI at distance $D$, $M$ points are sampled - typically on a grid - within the ROI. The optimization loss, $L_{ROI}$, is defined as the mean RMS spot size across these sampled points:

\begin{equation}
    L_{ROI} = \frac{1}{M} \sum_m^M r(P_m)
\end{equation}

\subsubsection{Joint optimization}

Although full-aberration correction can be achieved by tiling a grid of independently corrected ROIs, this approach does not efficiently exploit the information available across the entire image stack. Therefore, to jointly optimize aberration correction across the full FoV with a limited image budget $N$, we propose to optimize the RMS spot size grid across the stack collectively. 

We first sample a grid of $H \times W$ points over the entire FoV at a given depth. Ray-tracing is then performed for each of the $N$ different DPP wavefronts, and the corresponding RMS spot size is calculated, resulting in a stack of RMS spot size grid $\{r_n (i,j) \in R^{H \times W} | 1 \leq n \leq N\}$. Each element $r_n(i,j) = r_n(P_{ij})$ represents the RMS spot size of the sampled point $P_{ij}$, computed using the n-th DPP wavefront. The overall image quality after fusion depends on the minimal RMS spot size at each location across the stack, i.e. $r_{min}(i,j) = \min_n r_n(i,j)$. 
Therefore, to jointly optimize for all DPP patterns, we minimize the \textbf{grid stacking} loss.

\begin{equation}
    L_{gs} = \frac{1}{H \times W} \sum_{i,j} { r_{min}(i,j)}
\end{equation}

Essentially, the grid stacking loss ensures that for each wavefront, only the area where it performs best across the entire stack will be considered as its active region for optimization. However, optimization may yield some degenerate patterns where does not outperform all other waveforms anywhere in the image, thus having no active region at all. This usually happens when the budget of images increases. To alleviate it, we encourage these DPPs to optimize on the \textbf{hard regions}, using $r_{min} / \Bar{r}_{min}$ as a weight:

\begin{align}
    L_{hr} =&  \sum_{n'} \frac{1}{H \times W} \sum_{i,j} {r_{n'}(i,j)  \frac{r_{min}(i,j)}{\Bar{r}_{min}}} \\
    & n' \in \{n | \sum_{i,j} \mathds{1}\left(n,n^*\right) = 0 \}, \nonumber
\end{align}
where $\mathds{1}(\cdot)$ is the identity function, and $n^*(i,j) = argmin_{n} r_n(i,j)$ is the mask index for the sampled point, indicating which DPP wavefront provides the best image quality for this location. The overall loss for \textbf{joint optimization} is a sum of the two losses:

\begin{equation}
    L_{joint} = L_{gs} + L_{hr}
\end{equation}

The joint optimization scheme is validated using the same simulated optical system, as shown in Fig.~\ref{fig:psf_grid}. In the experiment, $32 \times 32$ grids are sampled across the full FoV to jointly optimize five wavefront patterns. The resulting optimized PSF grids exhibit compensatory patterns due to the gradients being masked differently by the index $n^*$ to optimize each DPP wavefront. As expected, the PSF grid displays radial symmetry, which the dynamic optimization scheme efficiently exploits to cover the entire FoV within a limited imaging budget.

\subsubsection{Efficiency under limited imaging budget}
\begin{figure}
    \centering
    \includegraphics[width=\linewidth]{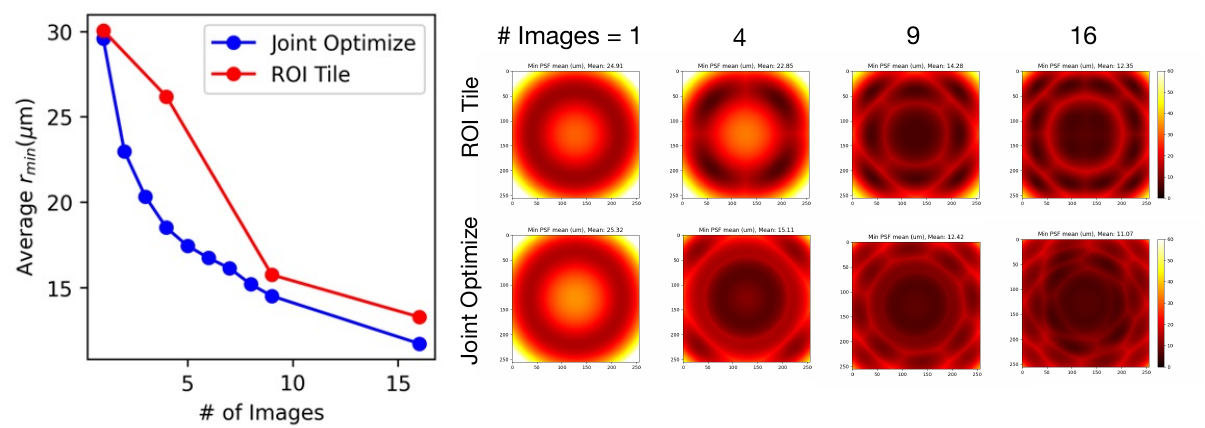}
    \caption{Comparison of the imaging efficiency between joint optimization and ROI tiling. }
    \label{fig:rPSF_vs_Nimg}
\end{figure}

The efficiency of using the imaging budget is analyized in Fig~\ref{fig:rPSF_vs_Nimg}. With a limited image count, we optimize DPP patterns and evaluate overall imaging quality using the average RMS spot size of the global PSF across the stack—i.e., the mean \(r_{min}\) (blue). ROI tiling requires selecting non-overlapping ROIs to cover the entire FoV; thus, image budgets of \(2\times2\), \(3\times3\), and \(4\times4\) are included for comparison (red). Joint optimization significantly improves image quality with as few as 3-5 images, whereas ROI tiling proves less efficient. Beyond 10 images, further improvements in image quality become marginal.

\section{DPP Control Model}

To control the DPP's deformation, earlier work used a linear model \cite{banerjee2018optofluidic,rajaeipour2021seventh}, which assumes the desired membrane shape is related to the square of the electrode voltages via a linear influence matrix. i.e.
\begin{equation}
    W = AV^2,
\end{equation}
where $W \in R^K$ represents the Zernike coefficients, and $V^2 \in R^{63}$ is the element-wise square of the electrode voltages. The matrix $A \in R^{K \times 63}$ is the influence matrix obtained through calibration. To calculate the voltage corresponding to a desired set of Zernike coefficients $W$, it is therefore modeled by solving a constrained optimization problem such that the voltage is within the physical limit $[0,V_{max}]$:
\begin{equation}
    \min || W- AV^2||_2^2 \quad s.t. 0<V<V_{max},
\end{equation}
where $V_{max}=270V$ is the maximum voltage for each electrode. This model is also implemented in the drivers of the DPP.

However, this model has two downsides: first, since the hardware capabilities limited by $V_{max}$, working in Zernike space makes it difficult to predict whether a desired shape is feasible on the device. Second, even if the target shape within the feasible range, we find that the linear model breaks down for larger amplitudes. To address this limitation, we propose a neural network-based control strategy that captures the non-linear behavior of the device and improve accuracy, thereby reducing the gap between physical implementation and simulation. Furthermore, by controlling the shape directly in voltage space, the feasibility of the desired wavefront correction is guaranteed.

In the following subsections, we propose our neural networks for DPP Control in Sec.~\ref{sec:sub_NN_models_for_DPP}, then the DPP's wavefront is measured using the experiment setup described in Sec.~\ref{sec:sub_wavefront_measurement} for neural network training in Sec.~\ref{sec:sub_dpp_train}, finally the control strategies is analyzed in Sec.~\ref{sec:ctrl-strategy}.

\subsection{Neural Networks for DPP Control}
\label{sec:sub_NN_models_for_DPP}
To calibrate the wavefront response with respect to the control voltage applied to the device, we experimentally apply different voltages and measure the resulting optical path differences using a customized wavefront sensor~\cite{wang2019quantitative}. Given a known aperture size, Zernike coefficients are fitted to reconstruct the measured wavefronts. By collecting a dataset of paired voltage inputs and corresponding Zernike coefficients, we train encoder-decoder neural networks to control the DPP.

\begin{figure}
    \centering
    \includegraphics[width=1.0\linewidth]{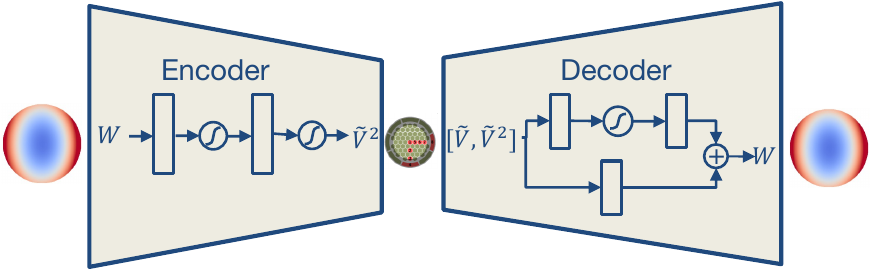}
    \caption{Neural networks for DPP control. Rectangular blocks stand
      for linear layers, activation functions are sigmoid
      functions. 
      The decoder’s residual design enables learning non-linear refinements based on the linear model's approximate estimation.
      }
    \label{fig:dpp_ctrl}
\end{figure}

The decoder predicts wavefront coefficients from control voltages using a dual-branch architecture: a linear branch to model overall linearity and a sigmoid-activated branch to capture residual nonlinearity. This residual learning strategy builds upon the linear model's approximate estimation~\cite{rajaeipour2021seventh}, requiring only nonlinear refinements. The input is extended to $[V, V^2]$, to facilitate pattern recognition. Conversely, the encoder, which predicts the desired control voltages from wavefront coefficients, is modeled as a non-linear mapping using a simple two-layer multilayer perceptron with a final sigmoid activation to constrain the output within the physical voltage limits. Voltage normalization by $V_{max}$ ($\tilde{V} = V/V_{max}$) is applied to ensures stable training.

\subsection{Wavefront Measurement}
\label{sec:sub_wavefront_measurement}
\begin{figure}
    \centering
    \includegraphics[width=\linewidth]{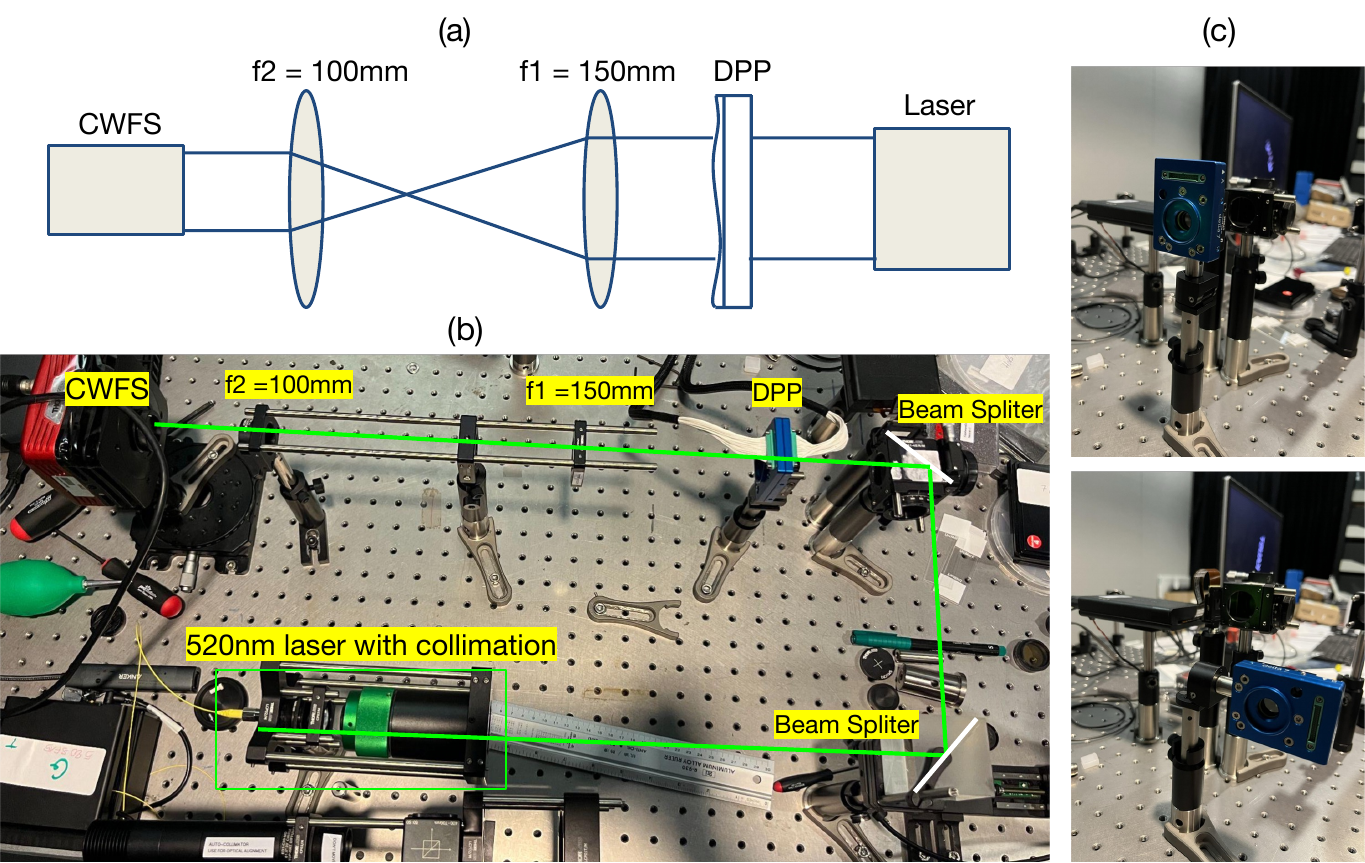}
    \caption{Wavefront Measurement Setup. (a) The optical path of the Measurement system. A 4F system with a spatial magnification of $-2/3$ relays the DPP device’s aperture plane to the wavefront sensor. (b) The realworld experiment setup. (c) The DPP is mounted on a flip-mount that can rotates 90 degrees to remove it from the optical path.}
    \label{CWFS_setup}
\end{figure}

A coded wavefront sensor (CWFS)~\cite{wang2019quantitative} is used to measure wavefront changes in the DPP. As shown in Fig~\ref{CWFS_setup}, a $4f$ system relays the DPP's aperture to the CWFS, with alignment ensured by an auto-collimator. Lenses with focal lengths of $150\,\mathrm{mm}$ and $100\,\mathrm{mm}$ provide a spatial magnification of $-2/3$, effectively projecting the full $10\,\mathrm{mm}$ DPP aperture onto most of the CWFS sensor area ($12.46 \times 6.7\,\mathrm{mm}$). Illumination is provided by a collimated $520\,\mathrm{nm}$ laser source. The DPP is mounted on a flip-mount that can rotate $90^\circ$, allowing it to be removed from the optical path to provide a collimated wavefront reference. The CWFS then measures the DPP's relative wavefront changes. From the measured relative wavefront, we fit the Zernike polynomials within the circular aperture region to get the corresponding coefficients. 

\subsection{Training of DPP Control Models}
\label{sec:sub_dpp_train}
\begin{figure}
    \centering
    \includegraphics[width=\linewidth]{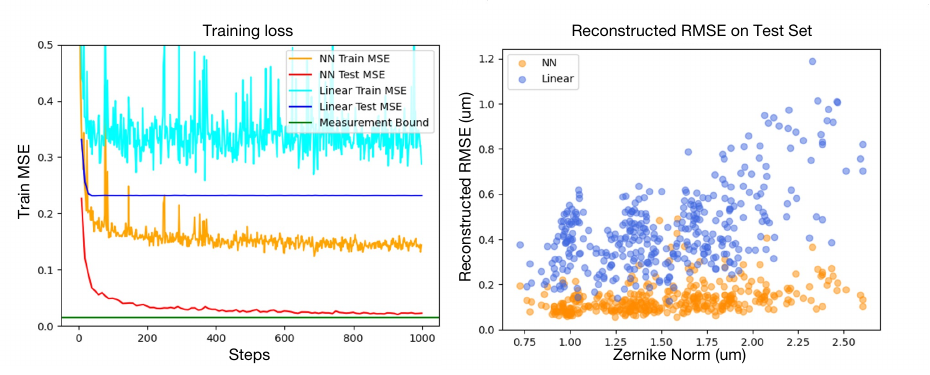}
    \caption{Decoder training loss and reconstruction error on the
      test set. (a) Training loss for the NN model is significant
      lower than for the linear model and approaches the measurement
      bound. (b) The NN model maintains lower reconstruction error for
      the full range of  deformations, while the linear model tends to performs worse for larger deformations (i.e. of larger Zernike norm). }
    \label{fig:dpp_ctrl_train}
\end{figure}

We captured 1800 voltage-Zernike coefficient pairs by randomly sampling Zernike coefficients within the device's operating range. We used $80\%$ of the data for training and the remaining $20\%$ for testing. More details on dataset collection are provided in the supplement. For comparison, a linear decoding model was trained alongside our neural-network (NN) model. To address measurement error and prevent overfitting, random noise with a standard devitation of 0.36 was added to the Zernike coefficients for data augmentation. As illustrated in Fig~\ref{fig:dpp_ctrl_train}, the neural network model achieves significantly lower training and testing loss than the linear model, approaching the measurement limit established by repeated CWFS measurements. On the test set, the linear model's reconstruction error (RMSE) increases with larger amplitudes, indicating poor performance at higher magnitudes, whereas the neural network maintains a more consistent error across magnitudes.

Unlike the linear model, which uses constrained least squares to compute the control voltage from the desired Zernike coefficients, we train an encoder neural network to predict the control voltage directly. This encoder achieves an RMS error of $2.56$ on the test set (in voltage control units).

\subsection{Control Strategies}
\label{sec:ctrl-strategy}
\begin{figure}
    \centering
    \includegraphics[width=\linewidth]{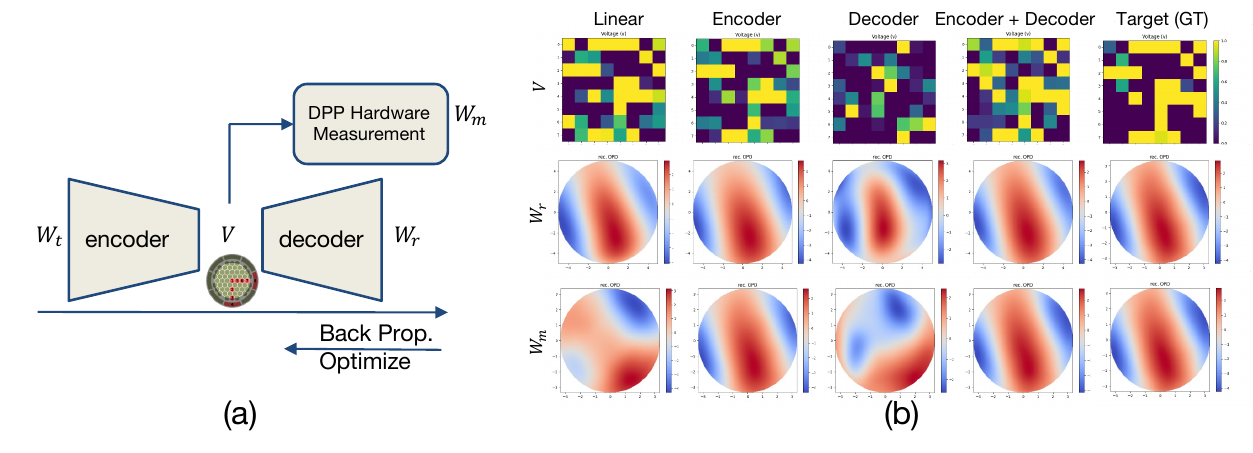}
    \caption{Comparison of different control strategies. (a) Given a
      target wavefront $W_t$, the control voltage $V$ is acquired by
      different control strategies (see text), and a prediction of the reconstructed wavefront $W_r$ is obtained from the model. To verify the effectiveness, the control signal $V$ is sent to the DPP device to obtained the measured wavefront $W_m$. (b) An example of different control strategies.  }
    \label{fig:dpp_ctrl_val}
\end{figure}

We compare different strategies for controlling the DPP to obtain the control voltage corresponding to a desired target Zernike Coefficient $W_t$, including the linear control strategy as a baseline. As shown in Fig.~\ref{fig:dpp_ctrl_val}, the "encoder" strategy directly predicts the control voltage $V$ via encoder neural network inference. The "decoder" strategy initializes $V$ as a zero tensor and then optimize it through back-propagation using the decoder, with $W_t$ as the target. The "encoder+decoder" strategy initializes $V$ as the predicted value from encoder and perform the following back-propagation through the decoder. For 10 known voltage-Zernike coefficient pairs in the test set, we applied each control strategy to obtain corresponding control voltages $V$. These voltages were then applied to the DPP device, and the resulting wavefronts $V_m$ were measured as described in Sec.~\ref{sec:sub_wavefront_measurement}.  

\begin{table}[]
    \centering
    \begin{tabular}{l|c c c c}
      MSE   & Linear    & Encoder       & Decoder  & Enc. + Dec.    \\
    \hline
    $V$     & 3.280      & \underline{2.560}          & 17.440    & \textbf{2.140}        \\
    $W_r$   & \underline{0.106}     & 0.119         & 0.217    & \textbf{0.002}        \\
    $W_m$   & 5.667      & \underline{0.166} & 1.160    & \textbf{0.103}                  \\
    $W_m$(h. o.) & 5.187 & \underline{0.099} & 1.098    & \textbf{0.013}
    \end{tabular}
    \caption{Quantative comparison of different control strategy. h.o. denotes high orders, meaning second order and above.}
    \label{tab:dpp_ctrl_val}
\end{table}

Table~\ref{tab:dpp_ctrl_val} presents the mean square error for various control strategies. The reconstructed voltage $V$ and model-predicted Zernike wavefront $W_r$ were evaluated across the entire test set (360 samples), while $V_m$ was measured for 10 target Zernike wavefronts. The encoder-decoder approach achieves the lowest control voltage error and consequently yields the most accurate wavefront measurement ($W_m$), particularly minimizing wavefront errors for higher-order Zernike aberrations (2nd order and above, excluding tilt). Figure~\ref{fig:dpp_ctrl_val} provides a qualitative example: the 63 control voltages are shown in an $8 \times 8$ grid, with the last entry set to zero. Note that this grid is a visualization and does not represent the actual physical layout of the electrodes.

\begin{figure}
    \centering
    \includegraphics[width=\linewidth]{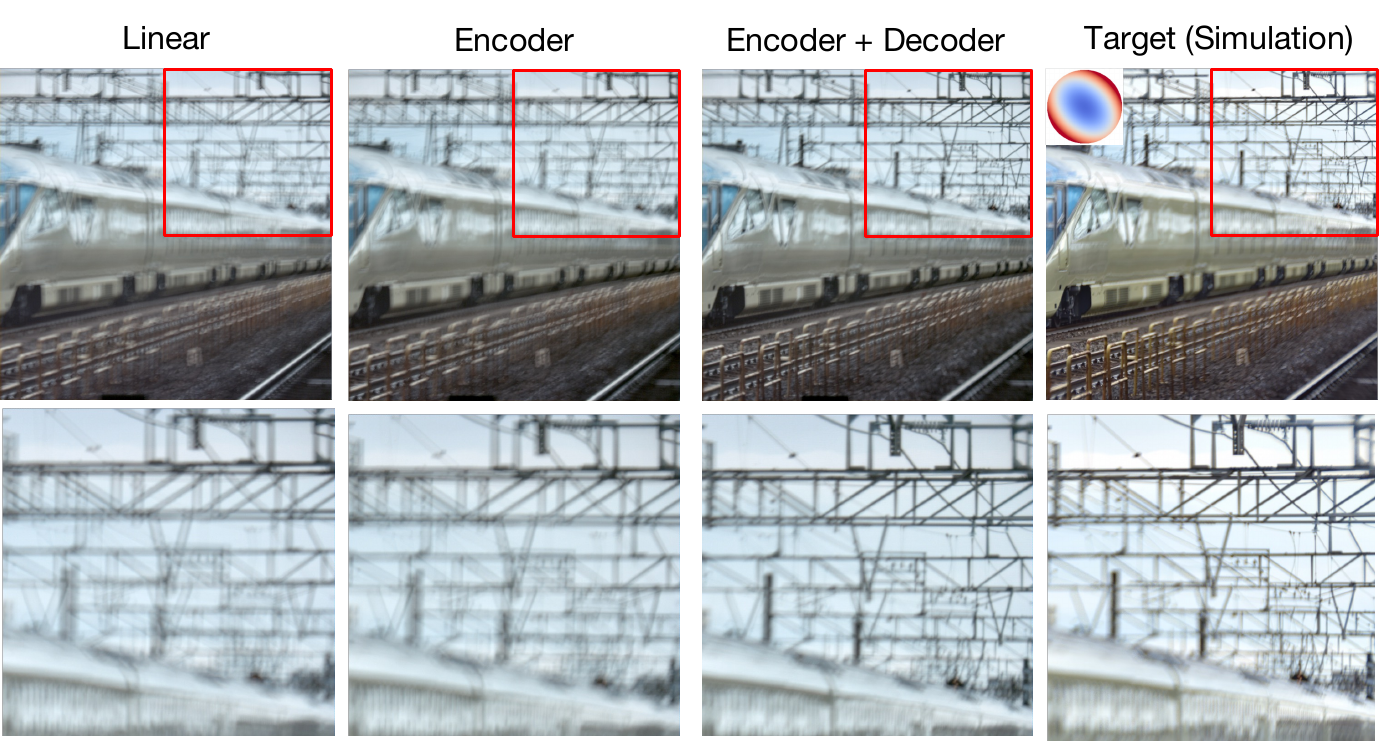}
    \caption{The captured images using different DPP control strategies. The encoder-decoder strategy best aligns with the target in simulation. }
    \label{fig:dpp-ctrl-img}
\end{figure}

We further verified the control strategies using the calibrated simulation model. As shown in Fig~\ref{fig:dpp-ctrl-img}, the target image is rendered in simulation using the desired wavefront pattern. Among all control strategies, the encoder+decoder method produces results that best align with the target image.

\section{Image Stacking}
\label{sec:image_stack}

To fuse the captured image stack, a straightforward approach involves pixel-wise fusion using the mask index $n^*$ obtained from the joint optimization process. However, real-world hardware imperfections - such as deviations in wavefront fidelity and calibration error - can lead to discrepancies between the simulated and actual system, degrading image quality when using the pre-optimized mask alone. Additionally, for extended depth-of-field applications, scene-dependent depth variations necessitate a flexible stacking algorithm.

While Fovea Stacking shares conceptual similarities with Focus Stacking, deep learning-based Focus Stacking methods are often trained on spatial-invariant, disk-pattern blur kernels, which offer limited improvement over traditional sharpness-based techniques. As a result, we adopt a conventional image fusion pipeline. We identify sharp regions by applying a Laplacian filter, followed by blurring. Each pixel in the fused image is computed as a weighted average across the image stack, with weights determined by sharpness. Further implementation details are provided in the supplement.
Experiments demonstrate that, despite its simplicity, this stacking method is effective in producing high-quality, aberration-corrected images.

\begin{figure}
\centering
    \includegraphics[width=\linewidth]{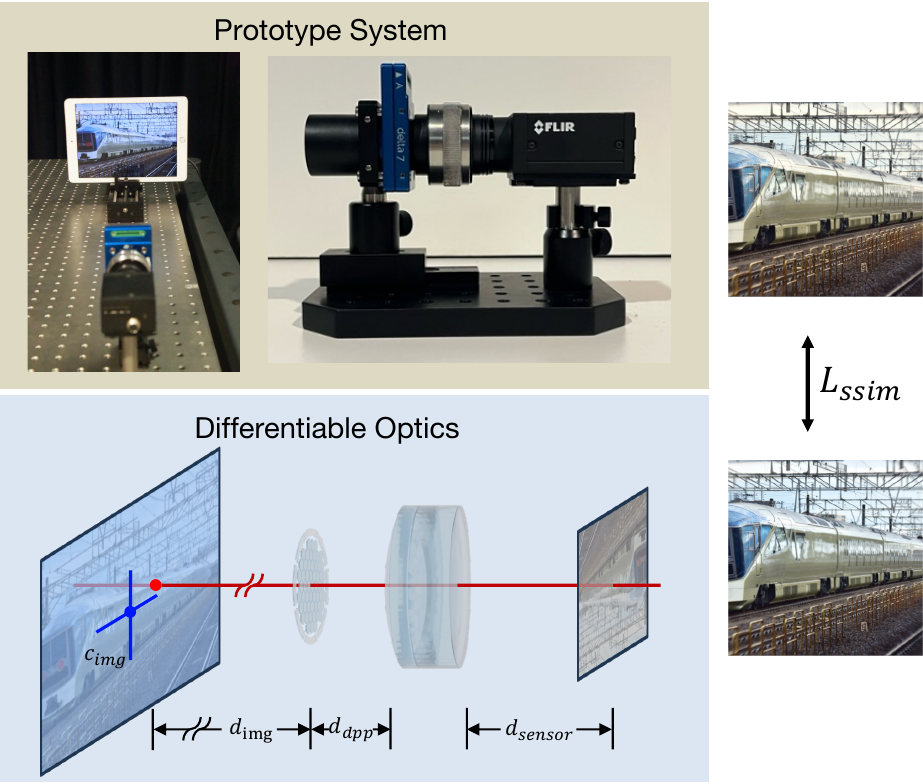}
    \caption{Prototype system and calibration. Our system comprises a DPP, an achromatic lens within a zoom enclosure for adjustable working distances, and a sensor. Assembly distances were calibrated by matching captured and simulated images.}
    \label{fig:cam_calib}
\end{figure}

\section{Prototype system and Calibration}
\label{sec:calibration}

Our prototype comprises a DPP (Phaseform Delta 7), a doublet lens, and an image sensor. As shown in Fig.~\ref{fig:cam_calib}, the lens is housed in a zooming enclosure (Thorlab SM1ZM) to adjust the back focus distance ($d_{sensor}$) for different working distances. A 50 mm achromatic doublet lens (Thorlab AC-254-050-A) was selected for its simplicity and inherent achromatic aberration compensation. The Bayer-pattern RGB sensor (FLIR GS3-U3-41C6C-C) features a pixel size of 5.5 $\mu$m $\times$ 5.5 $\mu$m with $2048 \times 2048$ resolution. The total length of the compact imaging system is 15 cm. Since the components are assembled using threaded lens tubes, the relative distances between optical elements require further calibration.

We calibrate the prototype system using an image-based method that leverages differentiable optics to optimize calibration parameters by matching captured images with simulated counterparts. This process addresses unmeasured distances within the camera setup: the distance between the DPP and the lens's front plane ($d_{dpp}$), and the distance between the lens's back plane and the sensor ($d_{sensor}$). The imaging model also incorporates the distance between the image plane and the DPP ($d_{img}$) and the shift between the image center and the intersection of the object plane with the optical axis ($c_{img}$). Given known screen and sensor dimensions, these parameters are jointly optimized. To ensure robust calibration, we capture images using multiple DPP patterns and employ a translation stage to precisely move the image plane.
Due to differences in display characteristics and sensor response, the pixel values of captured and simulated images may differ. Therefore, we use the structural similarity index measure (SSIM) to compare local image structure rather than pixel values.
All distances are initialized using rough physical measurements, and the image shift $c_{img}$ is calculated from a homography transform estimated by matching the captured image to the known screen pattern.

\section{Experiments}
In this section, we experimentally demonstrate the capabilities of the hardware prototype in three applications: aberration-corrected imaging via Fovea Stacking at far distances (Sec.~\ref{sec:sub_exp_imaging}), extended depth-of-field imaging via Fovea Stacking across various depths (Sec.~\ref{sec:sub_exp_EDOF}), and foveated object tracking during smooth pursuit movements (Sec.~\ref{sec:sub_exp_tracking}). Throughout the experiments, all images were captured in RAW and demosaiced using bilinear interpolation. We did not correct lens distortion because it remains consistent across images, and the DPP tilting terms were set to zero to prevent pixel shifts. 

\begin{figure*}
    \centering
    \includegraphics[width=\linewidth]{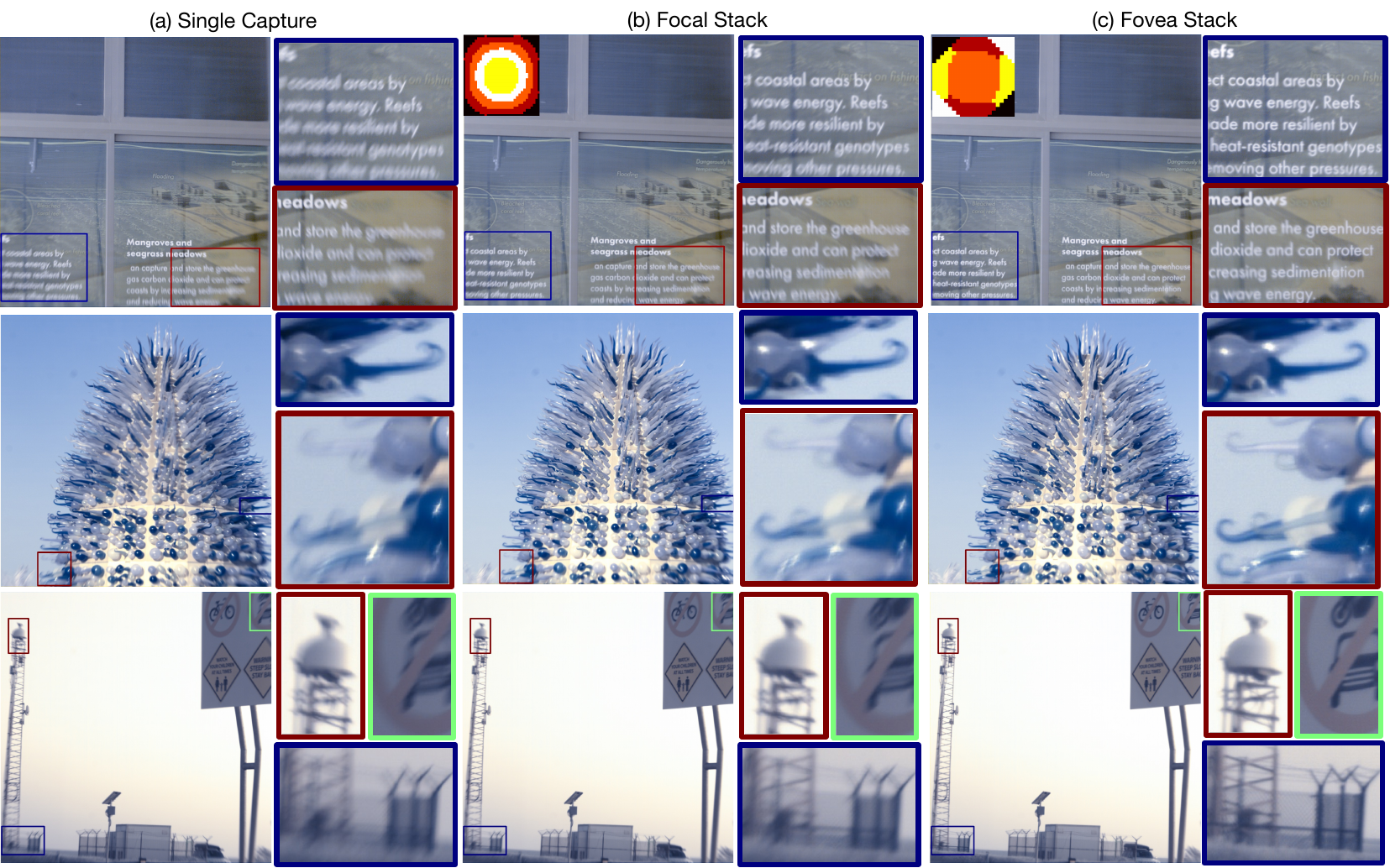}
    \caption{Fovea Stacking: (a) A single capture suffer from significant off-axis aberration. (b) Focus Stacking 5 images partially mitigates the curvature of field issue and thus improve the imaging quality, but doesn't eliminate the aberrations. (c) Fovea Stacking provides superior aberration correction. Mask index is visualized for stacking.}
    \label{exp:exp_fovea_stack}
\end{figure*}

\subsection{Fovea Stacking}
\label{sec:sub_exp_imaging}
For imaging beyond the hyperfocal distance, the sensor plane is re-positioned via zoom housing rotation. Five phase patterns are then jointly optimized at distance 60m as described in Sec.~\ref{sec:sub_joint_optimize} and used for image capture. For comparison, a focus stack with the same number of images was optimized using only defocus as a variable Zernike polynomial term. As shown in Fig.~\ref{exp:exp_fovea_stack}, single images are affected by significant off-axis aberrations (Fig.~\ref{exp:exp_fovea_stack}a). Although Focus Stacking can partially compensate for field curvature, it does not fully correct these aberrations (Fig.~\ref{exp:exp_fovea_stack}b). Fovea Stacking offers better aberration correction, as demonstrated by substantially improved resolution in fine structures and enhanced text readability in this comparison (Fig.~\ref{exp:exp_fovea_stack}c).  The stacking use the sharpness-based method described in Sec.~\ref{sec:image_stack}. 

\begin{figure}
    \centering
    \includegraphics[width=\linewidth]{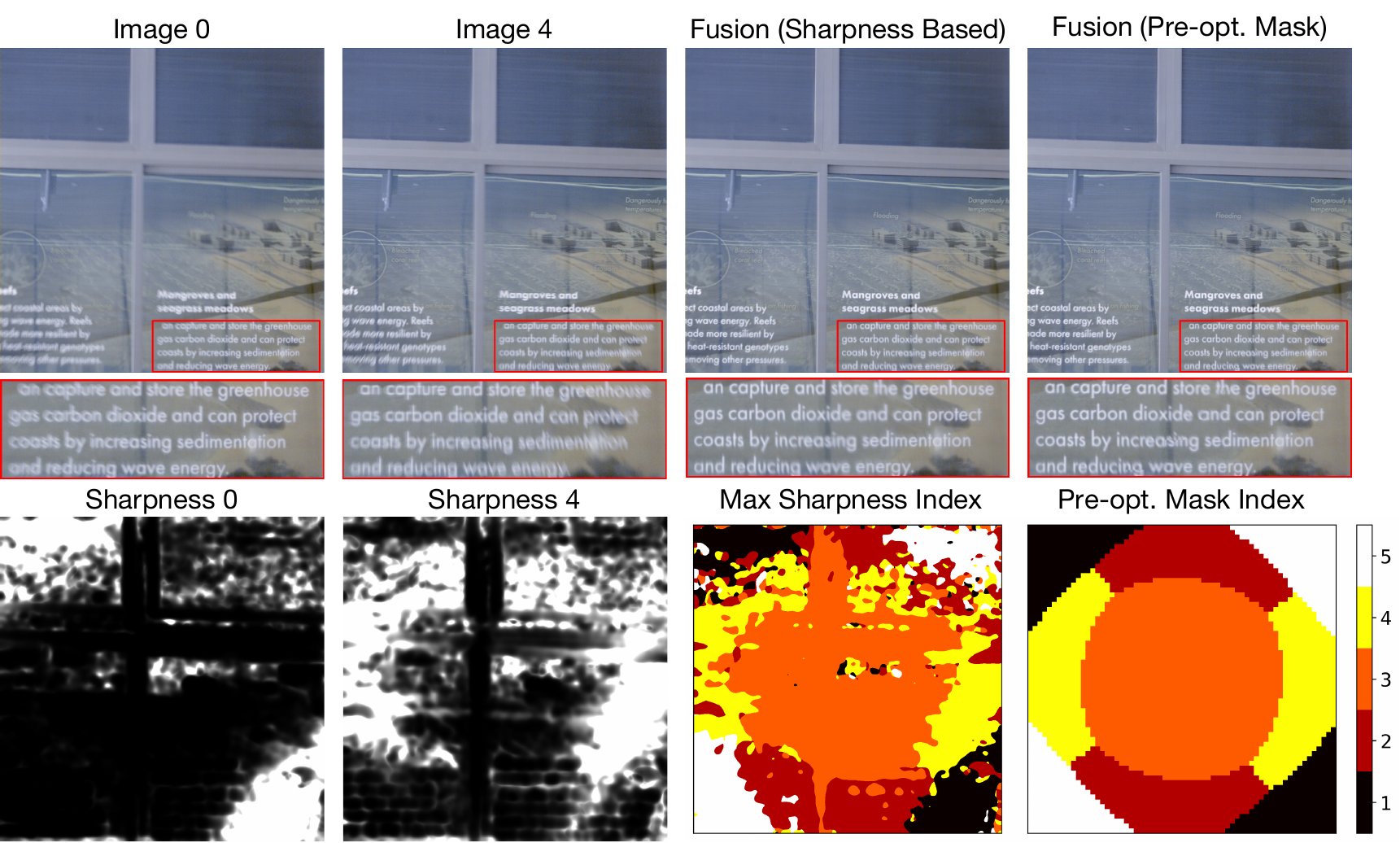}
    \caption{Comparison of Fusion by Image Sharpness vs. Pre-optimized Mask: The first two columns show the captured image $I_k$ and its corresponding sharpness map $S_k$. The third column presents the sharpness-based fusion result, along with the max sharpness index for each pixel. The final column displays fusion using a pre-optimized mask, which introduces noticeable artifacts along region boundaries. Artifacts are most apparent in the words "the," "increasing," and "reducing." }
    \label{fig:exp_fusion}
\end{figure}

Fig.~\ref{fig:exp_fusion} compares sharpness-based fusion (3rd column) with fusion using a pre-optimized mask (last column), which directly selects pixel values from the pre-optimized mask as described in Sec.~\ref{sec:sub_joint_optimize}. Real-world hardware imperfections introduce minor artifacts at image boundaries (most apparent in the words "the," "increasing," and "reducing."). In contrast, sharpness-based fusion mitigates these artifacts by providing smoother transitions at the boundaries. The sharpness map for each image aligns well with its pre-optimized mask, making it a reliable indicator for fusion.

\begin{figure}
    \centering
    \includegraphics[width=\linewidth]{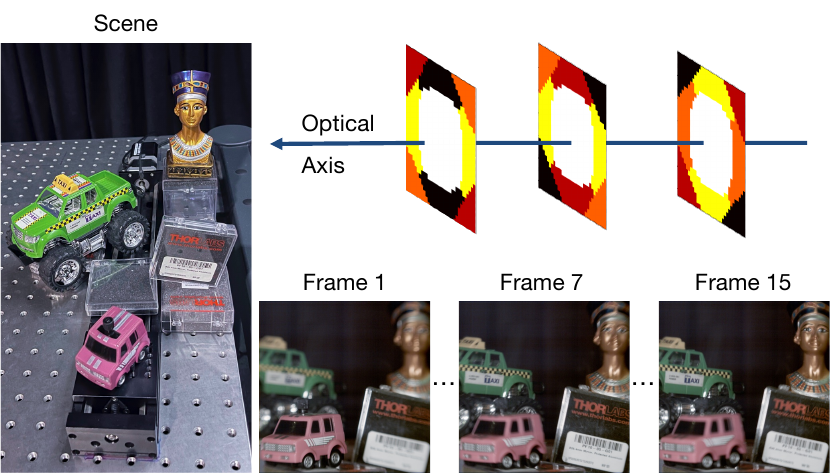}
    \caption{The Scene and scheme for extended depth of field. The scene composes 4 objects distributed within the range of 535 to 835 mm. 15 DPP patterns are optimized for three distances, each have 5 different patterns.}
    \label{fig:exp_fovea_EDOF_scheme}
\end{figure}

\begin{figure*}
    \centering
    \includegraphics[width=\linewidth]{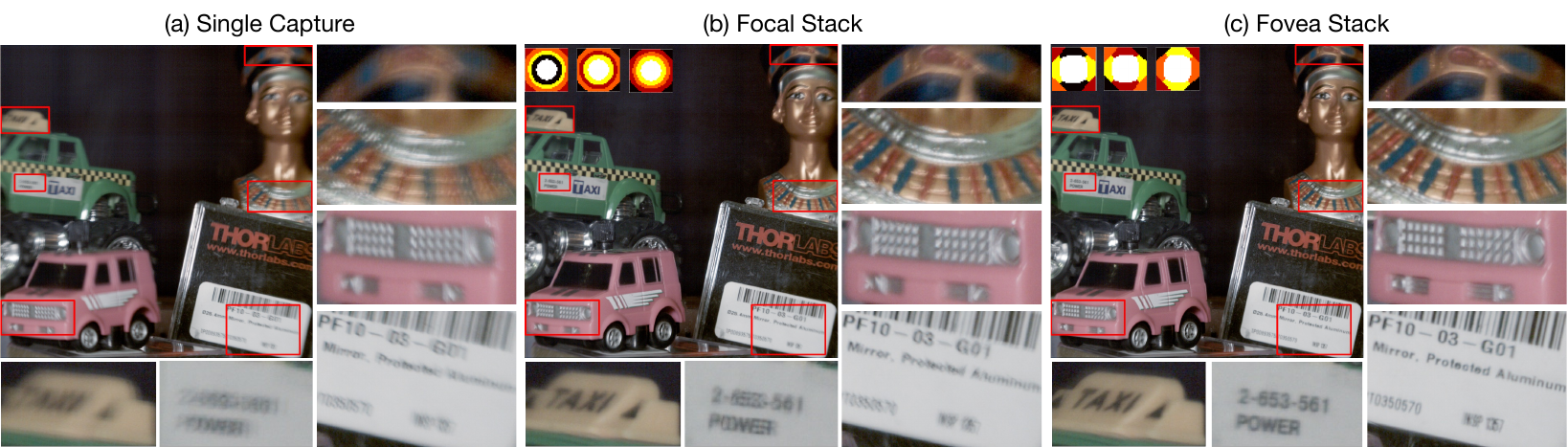}
    \caption{Fovea Stacking for Extended Depth of Field. (a) A single capture is limited by both out-of-focus blur and off-axis aberrations. (b) Focus Stacking reduces out-of-focus blur but does not correct off-axis aberrations. (c) Fovea Stacking corrects both, as seen in the improved sharpness and text clarity.}
    \label{fig:exp_fovea_EDOF}
\end{figure*}

\begin{figure*}
    \centering
    \includegraphics[width=0.9\linewidth]{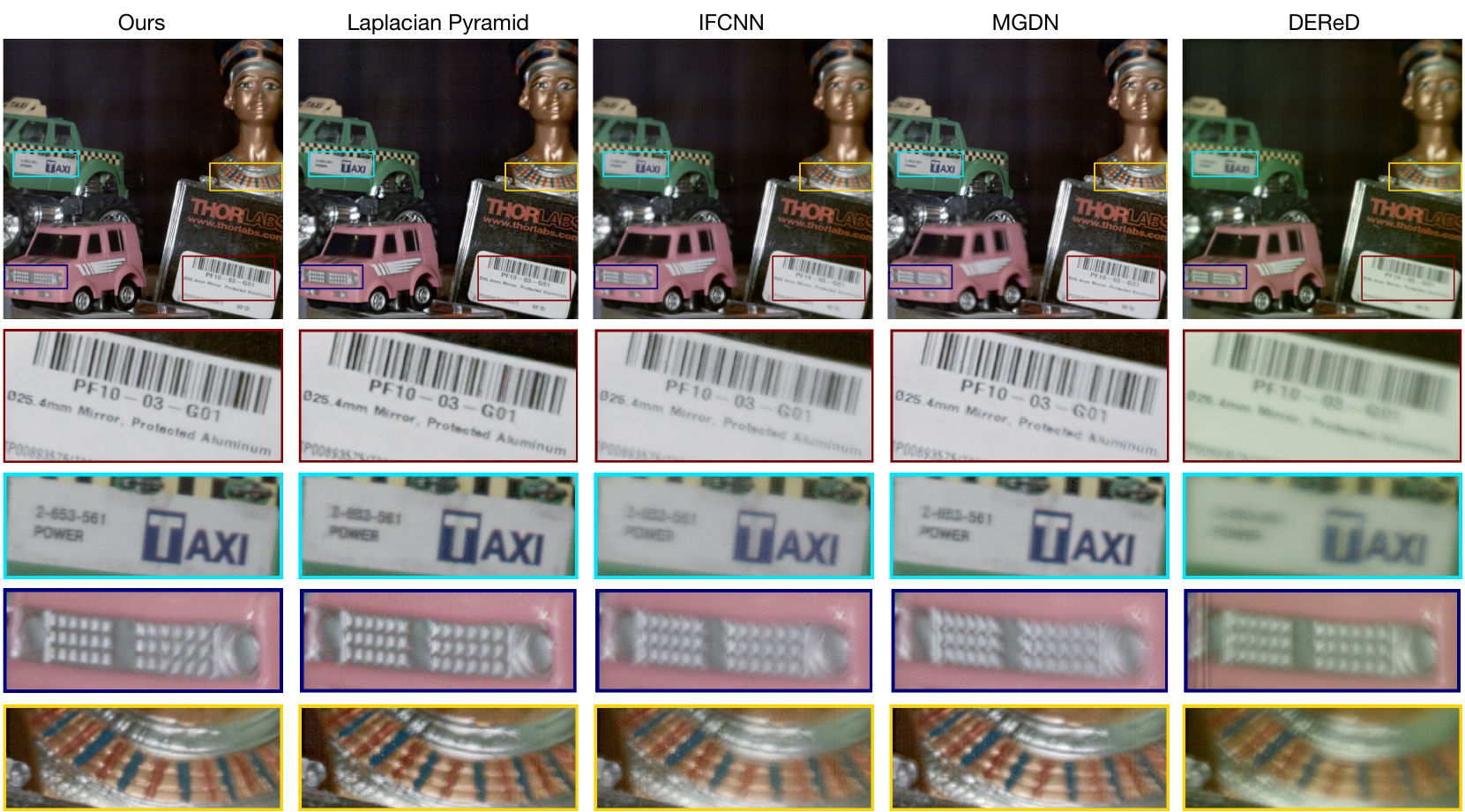}
    \caption{Comparison of different fusion methods for extended depth of field application. Pairwise fusion networks (IFCNN, MGDN) suffer from quality degradation due to sequential fusion. DEReD, although processing the entire stack, is prone to overfitting. Traditional Laplacian Pyramids offer improved sharpness, but still struggles in text readability and barcode resolution. Our method achieves the highest overall quality.}
    \label{fig:exp_fusion_compare}
\end{figure*}

\subsection{Fovea Stacking for Extended Depth of Field}
\label{sec:sub_exp_EDOF}
To achieve extended depth of field using Fovea Stacking (Fig.~\ref{fig:exp_fovea_EDOF_scheme}), we sampled three planes evenly in disparity space between 535 and 835 mm, optimizing five phases per plane for a total of 15 images. Four objects were placed at varying depths, with two car models positioned at angles to enhance the depth variation. For a fair comparison with the Focus Stacking method, we also optimized 15 phases, varying only the defocus term in the Zernike polynomials.

Fig.~\ref{fig:exp_fovea_EDOF} shows that a single capture (Fig.~\ref{fig:exp_fovea_EDOF}a) focused around 652 mm suffers from significant off-axis and out-of-focus aberrations. While Focus Stacking (Fig.~\ref{fig:exp_fovea_EDOF}b) mitigates out-of-focus aberrations, off-axis aberrations still limit resolution in peripheral areas. Fovea Stacking (Fig.~\ref{fig:exp_fovea_EDOF}c) corrects for both aberrations, and preserves finer details. 

For extended depth of field application, fusion with pre-optimized masks is unsuitable, as object depths vary arbitrarily. In Fig.~\ref{fig:exp_fusion_compare}, we compare our fusion method to Laplacian Pyramid~\cite{wang2011LaplacianPyramid}, IFCNN~\cite{zhang2020ifcnn}, MGDN~\cite{guan2023MGDN}, and DEReD~\cite{si2023DEReD}.  Pairwise fusion networks such as IFCNN and MGDN lose sharpness when sequentially fusing the entire image stack. Although DEReD processes the whole stack simultaneously, we observed that it overfits its training data and produces hue variations depending on the focus depth of each image. While the traditional Laplacian Pyramid method achieves comparable sharpness, it is less effective in text readability and barcode resolution. We attribute the lower quality of neural network-based methods to the introduction of different types of PSF in fovea imaging, compared to their original Focus Stacking domain where optics are typically well corrected for off-axis aberrations. In contrast, traditional sharpness-based and Laplacian-based methods are less sensitive to the blur kernel, resulting in better fusion quality.

\begin{figure*}
    \centering
    \includegraphics[width=\linewidth]{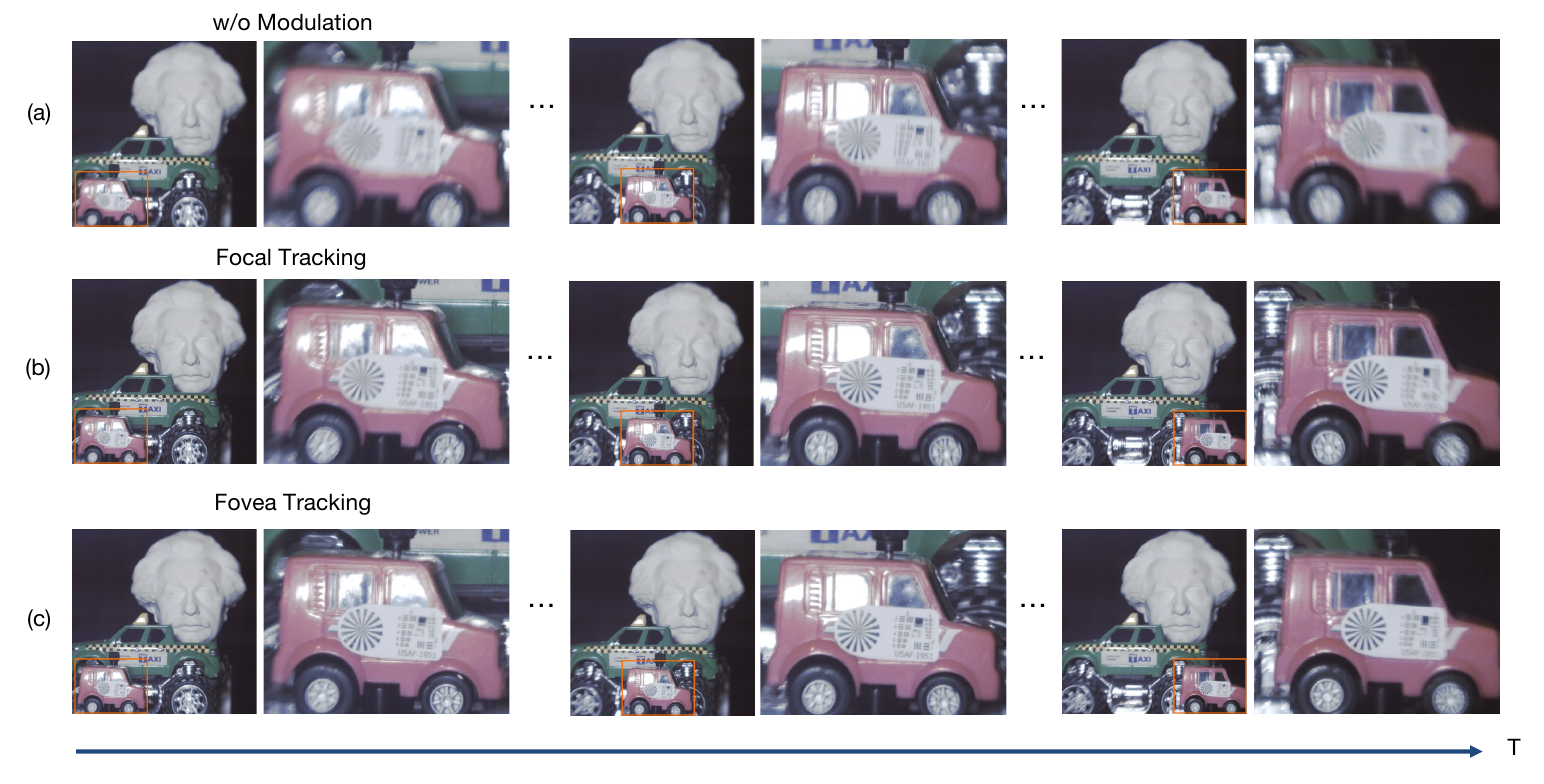}
    \caption{Foveated Imaging for Object Tracking. A car moves at 1 mm/s on a translation stage across the image. (a) Without modulation, the object appear blurred on the image edges. (b) With adaptive focus tracking, field curvature is corrected for better local imaging quality. (c) Fovea Stacking keeps the object sharp throughout the video. }
    \label{fig:exp_fovea_tracking}
\end{figure*}

\subsection{Foveated Imaging for Object Tracking}
\label{sec:sub_exp_tracking}

For surveillance video applications, where only a single object in the field of view is of interest. Our method tracks this object and ensures its corresponding area is imaged clearly. In our experiment, a pink car moves at 1 mm/s on a translation stage, positioned 652 mm from the camera. After selecting the object region in the first frame, we employ a tracking algorithm~\cite{bolme2010visual_track} to follow the target's bounding box throughout the sequence. For tracking application, the control voltage that locally corrects aberrations must be computed quickly to minimize latency. Because back-propagation optimization at each location is too slow for real-time use, we instead pre-optimize DPP deformation patterns on a 9 $\times$ 9 grid across the image and use linear interpolation to estimate the control voltage at any arbitrary location. Further analysis of the accuracy of interpolation control is provided in the supplement. For fovea tracking, the interpolated control voltage at the bounding box center is applied to the DPP device for image capture. For focus tracking, voltage is interpolated similarly, but using grids optimized only for defocus correction. Although DPP device have a response time of less than 55 ms, we wait for 200 ms before capturing the next frame to ensure stabilized deformation.

As shown in Fig.~\ref{fig:exp_fovea_tracking}, without DPP modulation, the optical system with a focus distance at 652 mm is blurred in off-axis regions (Fig.~\ref{fig:exp_fovea_tracking}a). Adaptive focus tracking corrects field curvature, improving local image quality (Fig.~\ref{fig:exp_fovea_tracking}b). Fovea Stacking maintains sharpness throughout the video (Fig.~\ref{fig:exp_fovea_tracking}c). Please also see the accompanying video.

\section{Conclusion}

We have presented a novel imaging approach that leverages a deformable phase plate (DPP) for dynamic, local aberration correction within a compact optical system. Our method introduces Fovea Stacking as a new paradigm for camera systems: by optimizing DPP wavefront control patterns using a differentiable optical model, we produce regionally corrected images across varying depths that can be stacked to form a high-quality  composite image with drastically reduced aberrations.

To efficiently cover the field of view with minimal saccadic movements, we proposed a joint optimization framework for DPP deformation patterns, enabling full aberration correction from as few as 3–5 stacked images. To address the non-linear behavior of DPPs, particularly for larger control signals, we developed a neural network-based control model that maps desired wavefronts to actuation patterns, thereby bridging the gap between simulation and real-world performance.

Our experiments demonstrate the system’s capabilities in aberration-corrected imaging and extended depth-of-field imaging. Comparing to focus adjustment (either by shifting the lens or with a liquid tunable lens), the DPP's dynamic free-shape deformation offers superior flexibility for enhancing image quality across diverse applications. Analysis demonstrates the robustness of the proposed sharpness-based fusion method, while the neural network-based approach fails to adapt to the blur kernel in Fovea Stacking. By integrating foveated imaging with object detection or eye tracking, we enable smooth pursuit of moving targets, dynamically adjusting the imaging focus to maintain the object within the fovea region—opening new possibilities for applications such as surveillance and foveated virtual reality displays.

These contributions are validated on a functional hardware prototype compact enough for use beyond controlled lab environments. As DPP devices evolve toward greater miniaturization, this method could allow their integration into mobile devices--—similar to liquid tunable lenses—--which is not feasible with other reflective wavefront modulators. We believe this work highlights how evolving dynamically tunable optical components can help simplify optical systems and redefine the boundary between optics and computing in future imaging devices.

The system's ability to correct the aberration is limited by the maximum achievable deformation of the hardware, which could be overcome by cascading multiple thin DPPs~\cite{rajaeipour2020cascading}. Promising future directions include real-time dynamic fovea sweeping and adaptive saccadic imaging guided by scene content. Since foveated imaging projects a 3D scene into multiple 2D views, finer 3D scene details can be reconstructed from foveated images without moving the camera. For applications like extended depth of field imaging, optimal imaging parameters can be determined based on current 3D world estimates, enabling adaptive saccadic imaging.

\bibliographystyle{ACM-Reference-Format}
\bibliography{bibliography}

\end{document}